\documentclass{article}

\usepackage{PRIMEarxiv}

\usepackage[utf8]{inputenc} 
\usepackage[T1]{fontenc}    
\usepackage{hyperref}       
\usepackage{url}            
\usepackage{booktabs}       
\usepackage{amsfonts}       
\usepackage{nicefrac}       
\usepackage{microtype}      
\usepackage{lipsum}
\usepackage{fancyhdr}       
\usepackage{graphicx}       
\graphicspath{{media/}}     
\usepackage{amsmath}

\title{Effects of Parametric and Non-Parametric Methods on High Dimensional Sparse Matrix Representations}

\author{
  Sayali Tambe \\
  Master of Engineering \\
  Vidyalankar Institute of Technology \\
  \texttt{sayalitambe2004@gmail.com} \\
   \And
  Raunak Joshi \\
  Student Mentor \\
  University of Mumbai \\
  \texttt{raunakjoshi.m@gmail.com} \\
   \And
   Abhishek Gupta \\
   Developer \\
   University of Mumbai \\
   \texttt{abhishek.gupta20001@gmail.com} \\
   \And
   Nandan Kanvinde \\
   Developer \\
   University of Mumbai \\
   \texttt{kanvindenandan81@gmail.com} \\
   \And
   Dr. Vidya Chitre \\
   Assistant Professor \\
   Vidyalankar Institute of Technology \\
   \texttt{vidya.chitre@vit.edu.in} \\
}

\begin{document}
\maketitle

\begin{abstract}
The semantics are derived from textual data that provide representations for Machine Learning algorithms. These representations are interpretable form of high dimensional sparse matrix that are given as an input to the machine learning algorithms. Since learning methods are broadly classified as parametric and non-parametric learning methods, in this paper we provide the effects of these type of algorithms on the high dimensional sparse matrix representations. In order to derive the representations from the text data, we have considered TF-IDF representation with valid reason in the paper. We have formed representations of 50, 100, 500, 1000 and 5000 dimensions respectively over which we have performed classification using Linear Discriminant Analysis and Naive Bayes as parametric learning method, Decision Tree and Support Vector Machines as non-parametric learning method. We have later provided the metrics on every single dimension of the representation and effect of every single algorithm detailed in this paper.
\end{abstract}


\section{Introduction}
Allocation of text oriented data is done on regular basis and it can be used for deriving semantics for which Natural Language Processing \cite{khurana2017natural} is used. Many primordial techniques are used for processing the text and semantics are derived with Machine Learning. Parametric and Non-Parametric \cite{7321316} are two major types of Machine Learning methodologies. For any text oriented task to work with, formatting and processing of the text is essential for machine learning algorithm. This essentially includes using stop-words \cite{sarica2021stopwords} and formatting into some representation \cite{naseem2021comprehensive} state. Representations is a technique used for structuring data in numerical format which derive representations for the sentences, from which semantics can be derived in the latter part. The representations of the text can be derived using techniques like one-hot encoding \cite{Hancock2020SurveyOC}, bag-of-words \cite{Zhang2010UnderstandingBM}, TF-IDF \cite{Qaiser2018TextMU}. These representation techniques will be explained in detail in Section \ref{sec:methodology}. Representations alone cannot be sufficient, in order to derive semantics from text, one requires machine or deep learning techniques. The machine learning techniques we have used for deriving semantics from the representations in this paper are parametric and non-parametric. In the parametric learning methods, we have focused on Linear Discriminant Analysis \cite{salford52074,ghojogh2019fisher,dorfer2015deep, gupta2022discriminant}, Naive Bayes \cite{Watson2001AnES, Kaviani2017ShortSO, 6045027} and in non-parametric methods we have focused on Decision Tree \cite{10.1023/A:1022643204877} and Support Vector Machines \cite{Cristianini2008, 708428}. These are most commonly used algorithms in the area of Machine Learning and these will be explained in detail of Section \ref{sec:methodology}.

\section{Methodology}
\label{sec:methodology}

\subsection{Word Representation}
The primary step when dealing with a NLP problem is processing of the text oriented data in word representations. These representations are interpretable by the Machine Learning algorithms. Bag-of-Words \cite{Zhang2010UnderstandingBM} is a very lower level word representation, as it only capable of holding the occurrence of the data. The representation lies in integer format and is very straightforward. It makes a vocabulary of known words. This parameter is predefined and maps the occurrence of every word that is found from the corpus. This is the reason it is called bag-of-words, as it bags all the words in some dimensional matrix. The derivable semantics from such a representation for machine learning algorithm is difficult as there is no mapping of words with syntactical meaning. Everything is just maintained by a count of the words from document from known corpus. The calculable complexities arise as the sparsity of matrix can increase with the size of vocabulary. Context is a very different issue, but regarding the issues of space complexity when compared with the syntactical interpretability is very inefficient. To tackle this problem Term Frequency and Inverse Document Frequency abbreviated as TF-IDF \cite{Qaiser2018TextMU}. The terminology of each word holds a very high meaning in this case. Term is nothing but every word, Document is nothing but set of words as words make up a document. The process starts by vectorizing the documents into vocabulary. This vocabulary is entire corpus for word representations. As the abbreviation is named, one must first consider TF, then DF, then inverse the DF and finally multiply both the terms. The formula of term frequency is represented by

\begin{equation}
    TF(T, D) = \frac{D(|T|)}{D[N(T)]}
\end{equation}

where TF is Term Frequency, that considers parameters as Term and Document. The value is fraction of occurenece of term in entire document by total number of words in entire document. The next thing to consider is document frequency. The formula for it is given by

\begin{equation}
    DF(T) = D[C(T)]
\end{equation}

where DF is document frequency where T is its parameter and it is calculated on the basis of count of all the terms in the entire document. Considering this document frequency, one can calculate the inverse document frequency. The formula for which can be given as

\begin{equation}
    IDF(T) = \log\frac{N(D)}{1+DF(T)}
\end{equation}

where IDF is inverse document frequency that takes term as its parameter. The $N(D)$ term is the count of entire corpus, and it is fractioned with $1+DF(T)$. This can be now used to obtain entire term of TF-IDF. The formula for which is given by

\begin{equation}
    TF-IDF(T, D) = TF(T, D)*IDF(T)
\end{equation}

where both, term frequency and inverse document frequency are multiplied with each other. Since the inverse document frequency has a logarithmic constant included in it, squashing of the values between [0,1] is performed. This gives the values in floating number values between 0 and 1 and is better than stationary integer values generated in bag of words. These representations will produce better semantics deriving as compared to the bag of words. The association of the words in sentence in done in a detailed manner which is better for any learning method. In our problem we have used TF-IDF dimensions of matrix sparsity of over 50, 100, 500, 1000, 5000 feature set. This will help the learning methods to check performance on every possible algorithm to find the most suitable algorithm.

\subsection{Parametric Methods}
After getting set of representations from the TF-IDF, we first take parametric learning methods into consideration. The algorithms we specifically focus on are Linear Discriminant Analysis and Naive Bayes. The Linear Discriminant Analysis is a supervised dimensionality reduction algorithm that works in a linear fashion and is parametric. The parametric methods calculate their specified parameters before making predictions in the training phase of the algorithm. The prediction function uses the parameters calculated and these parameters are stationary. For Linear Discriminant Analysis abbreviated as LDA, consideration of $d$ dimensional data points is done, where $W$ is considered as a unit vector. When samples are considered as $x(n)$ the feature space is denoted by $W^T.x(n)$ for the projection process. The means of the classes are denoted by $m_i$ and after projection of the classes they become $W^T.m_i$ respectively. In the later part, the estimates of co-variance matrix are used which are known as Scatter Matrix \cite{xu2018weighted}. This is calculated by $\sigma*N$ where $\sigma$ is sample variance and $N$ is number of samples. So basically 2 types of scatter matrices are used. Between Class Scatter denoted by $S_b$ and Within Class Scatter denoted by $S_w$. This finally is represented by a formula as

\begin{equation}
    LDA = \frac{(M_1-M_2)^2}{(S^2_1+S^2_2)}
\end{equation}

where numerator is $S_b$ and denominator is $S_w$ where maximizing both the terms is essential. For maximizing the numerator the approach is given as follows

\begin{flalign}
    S_b = (M_1-M_2)^2 \nonumber \\
    S_b = (W^T.m_1-W^T.m_2).(W^T.m_1-W^T.m_2)^T \nonumber \\
    S_b = W^T.(m_1-m_2).(m_1-m_2)^T.W \nonumber
\end{flalign}

which gives the finally equation as

\begin{equation}
    S_b = W^T.S_b.W
\end{equation}

similarly within class scatter matrix can also be calculated by the formula as

\begin{equation}
    S^2_i = W^T.\left(\sum(x(n)-m_i).(x(n)-m_i)^T\right)W \nonumber
\end{equation}

which further gives within scatter matrix as

\begin{equation}
    S^2_1 = W^T.S_i.W
\end{equation}

Now these scatter matrices are differentiated that yield eigenvalues and eigenvectors \cite{denton2022eigenvectors} which gives a full ranked matrix.

Another parametric algorithm we have used is Naive Bayes, which works with probabilities in different contexts for learning. The Naive Bayes use the primordial Bayes Theorem for calculating the values. It takes the prior probabilities into consideration and calculates the posterior probabilities. The formula is given by

\begin{equation}
    P(c|x) = \frac{P(x|c)*P(c)}{P(x)}
\end{equation}

where $P(c|x)$ is the posterior probability, $P(x|c)$ is the Likelihood, $P(c)$ is the Class Prior Probability and $P(x)$ is the Predictor Prior Probability. $P(c|x)=P(x_1|c)*P(x_2|c)*\dots*P(x_n|c)*P(c)$ is the derived equation in an expansive form.

\subsection{Non-Parametric Methods}
These methods do not operate with stationary parameters derived from the data. The learning procedures are very different and the algorithms that we have considered for our problem are Decision Trees and Support Vector Machines. The Decision Tree are used for Classification and Regression techniques and are derived from the concept of CART algorithm. It is a supervised learning approach and works with node structure. The feature importance is of higher value in the Decision Tree and associations can be easily interpreted since the it is highly outcome oriented. Since this algorithm is outcome oriented, the need of standard scaling the data is eliminated. The system of the algorithm works with splitting the data according to the outcomes and this approach can term it is a Greedy Learning Approach. Other algorithm we considered is Support Vector Machines which is a supervised non-parametric learning method that uses a term known as support vector which are nothing but points close to hyperplane used for classification operation. The hyperplane is a line passed from set of datapoints which divides the points equidistantly. This hyperplane has a margin for best possible identification of the support vectors. Closer the points that touch the margin, make it to the classified label group. The equation for hyperplane can be given as

\begin{equation}
    w^T.x+b=0
\end{equation}

where $w$ is normal for vector and $b$ is an offset for hyperplane. This is where the decision rule comes into practice, where the algorithm decides to consider one aspect of equation to be positive and other negative. This is used for calculation of the maximal margin which is denoted by $d$. These margin lines are closer to the points of data, which are regarded as support vectors. For this the equations become $w^T.x+b=-1$ and $w^T.x+b=1$. These equations represent the maximal margin lines. These further extends into kernelization for support vector machines. These kernels give you levearage to interact with an extensive form of data in higher dimensions, especially non-linear. Radial Basis Filtering abbreviated as RBF \cite{thurnhofer2020radial} is of the non-linear nature kernel which we have used in our implementation. It helps one, create a linear decision boundary in even very high dimensional space. The formula for it is given by

\begin{equation}
    RBF = e^y \nonumber
\end{equation}
where y is denoted by
\begin{equation}
    y = \frac{-||x_1-x_2||^2}{2\sigma^2}
\end{equation}

where function in the numerator is Euclidean Distance \cite{Liberti2014EuclideanDG} between 2 points and $\sigma$ is the variance for kernel, which is considered as a hyperparameter. 

\subsection{Data}
The problem being not very data specific and works on proving a point, the room for selection of the data was very broad for us. We specifically searched for a data that had certain number of classes which can clearly distinguish the text. Our main type of data for such a problem was Text Classification. The data we decided to work with is Cyberbullying Detection \cite{9378065} from Tweets. The freedom offered for expressing opinions is not bounded by any constraints and this is where the cyberbullying started as a retaliation of some opinions that are not acceptable by certain societies. This became a medium of exploitation and was used extensively. This can be detected in early stages and leveraged using NLP as the data allocated on social media is textual. The text classification is done for this purpose as the detection of type of cyberbullying was main concern of the data. This data is collected from Twitter which gives tweets of cyberbullying in context of Religion, Ethnicity, Age and Gender. This purely is a text classification problem and was completely adjustable with the problem we are trying to solve in this paper.

\section{Results}

\subsection{Precision}
The Precision \cite{powers2020evaluation} is the first metric that gives the value representation of the positive predictions yielded from overall positive predictions. This is based on the elements of confusion matrix \cite{Ting2017} viz. True Positives, True Negatives, False Positives and False Negatives. This is calculated by the formula

\begin{equation}
    Precision = \frac{TP}{TP+FP}
\end{equation}

For which we have calculated precision for all the algorithms, where we have 2 parametric and 2 non-parametric learning methods.

\begin{table}[htbp]
 \caption{Precision for 50 Dimensions}
  \centering
  \begin{tabular}{lll}
    \toprule
    Algorithm     & Macro Average & Weighted Average \\
    \midrule
    LDA & 89\% & 90\% \\
    Naive Bayes & 79\% & 80\% \\
    Decision Tree & 89\% & 90\% \\
    SVM & 90\% & 91\% \\
    \bottomrule
  \end{tabular}
  \label{tab:a}
\end{table}

The Table \ref{tab:a} is the precision score for 50 dimensions of the sparsity matrix representation for TF-IDF. The precision is considering Macro Average and Weighted Average methods of calculation. Macro is nothing but unweighted average is considered while weighted average is considering mean of every single class before calculation. This similar thing is applied to the 100, 500, 1000, 5000 representations of sparse matrix.

\begin{table}[htbp]
 \caption{Macro Precision for 100, 500, 1000, 5000 Dimensions}
  \centering
  \begin{tabular}{lllll}
    \toprule
    Dimensions & LDA & Naive Bayes & Decision Tree & SVM \\
    \midrule
    100 & 90\% & 83\% & 89\% & 92\% \\
    500 & 91\% & 85\% & 91\% & 93\% \\
    1000 & 92\% & 86\% & 91\% & 93\% \\
    5000 & 92\% & 86\% & 92\% & 93\% \\
    \bottomrule
  \end{tabular}
  \label{tab:b}
\end{table}

From Table \ref{tab:b} we can infer that higher the dimensions for sparsity matrix, the precision does not affect much. The SVM is able to generate a good result. Similar precision can be calculated for the Weighted Precision.

\begin{table}[htbp]
 \caption{Weighted Precision for 100, 500, 1000, 5000 Dimensions}
  \centering
  \begin{tabular}{lllll}
    \toprule
    Dimensions & LDA & Naive Bayes & Decision Tree & SVM \\
    \midrule
    100 & 91\% & 83\% & 90\% & 92\% \\
    500 & 92\% & 85\% & 91\% & 93\% \\
    1000 & 92\% & 86\% & 91\% & 93\% \\
    5000 & 92\% & 86\% & 92\% & 94\% \\
    \bottomrule
  \end{tabular}
  \label{tab:c}
\end{table}

The Table \ref{tab:c} clearly shows the minute differences between algorithms. A very small fluctuation can be seen in all algorithms except Naive Bayes.

\subsection{Recall}
The Recall \cite{powers2020evaluation} is another metric that gives the retrieve true samples from all the overall samples. The equation is given by

\begin{equation}
    Recall = \frac{TP}{TP+FN}
\end{equation}

Considering recall metrics let us calculate recall for 50, 100, 500, 1000, 5000 sparsity matrix representation over the algorithms used.

\begin{table}[htbp]
 \caption{Macro Average Recall for all dimensions}
  \centering
  \begin{tabular}{lllll}
    \toprule
    Dimensions & LDA & Naive Bayes & Decision Tree & SVM \\
    \midrule
    50 & 88\% & 79\% & 88\% & 90\% \\
    100 & 89\% & 82\% & 89\% & 91\% \\
    500 & 91\% & 84\% & 91\% & 93\% \\
    1000 & 91\% & 85\% & 91\% & 93\% \\
    5000 & 91\% & 85\% & 92\% & 93\% \\
    \bottomrule
  \end{tabular}
  \label{tab:d}
\end{table}

The Table \ref{tab:d} gives the Recall with Macro Averaging system. It can be derived that Naive Bayes has less recall than all the other algorithms. But still inferring anything is not a legitimate practice as one metric in machine learning cannot derive what is best. Similarly we will now use the Weighted Average Recall for all the dimensional sparsity matrix representation.

\begin{table}[htbp]
 \caption{Weighted Average Recall for all dimensions}
  \centering
  \begin{tabular}{lllll}
    \toprule
    Dimensions & LDA & Naive Bayes & Decision Tree & SVM \\
    \midrule
    50 & 88\% & 80\% & 89\% & 90\% \\
    100 & 89\% & 84\% & 89\% & 91\% \\
    500 & 91\% & 85\% & 91\% & 93\% \\
    1000 & 91\% & 86\% & 91\% & 93\% \\
    5000 & 92\% & 87\% & 92\% & 93\% \\
    \bottomrule
  \end{tabular}
  \label{tab:e}
\end{table}

Deriving a lot of information from any table is not of utter importance but deriving the subtle changes is the key. With a keen observation we can infer from Table \ref{tab:e} that the Naive Bayes does not get stagnant and increases Recall with the improvement in the dimensions of sparsity matrix for representations.

\subsection{$F_1$-Score}
This is the metric that can give you more insights as compared to the Precision and Recall. It uses both of them for calculations of the $F_1$-Score \cite{powers2020evaluation}. It is a metric of accuracy and represented by the formula

\begin{equation}
    F_1 = 2*\frac{P*R}{P+R}
\end{equation}

The Table \ref{tab:f} gives a detailed values for 50, 100, 500, 1000, 5000 dimensions of sparsity matrix for representations of TF-IDF.

\begin{table}[htbp]
 \caption{$F_1$-Score for all dimensions}
  \centering
  \begin{tabular}{lllll}
    \toprule
    Dimensions & LDA & Naive Bayes & Decision Tree & SVM \\
    \midrule
    50 & 88\% & 80\% & 89\% & 90\% \\
    100 & 89\% & 84\% & 89\% & 91\% \\
    500 & 91\% & 85\% & 91\% & 93\% \\
    1000 & 91\% & 86\% & 91\% & 93\% \\
    5000 & 92\% & 87\% & 92\% & 93\% \\
    \bottomrule
  \end{tabular}
  \label{tab:f}
\end{table}

The $F_1$-Score is a considerable metric that gives very good insights of the performing algorithms. The $F_1$-Score infers that all the algorithms perform very efficiently when the representations are very high. These representations practically determine the vocabulary of the entire corpus. The algorithms whether parametric or non-parametric, the difference is very negligible.

\section{Conclusion}
The semantics derived in any Natural Language Processing problem can be performed using Machine Learning where the consideration lies which prevail over the other, the parametric learning methods or non-parametric learning methods. In order to test the prowess of the algorithms for NLP problem, processing of the text oriented data into representations interpretable for the Machine Learning algorithms is required. For such a scenario we decided to consider TF-IDF representations over Bag-of-Words as it provides more intricate details about the data for deriving semantics. The TF-IDF gives a very high dimensional sparse matrix of representations that we used as an input for our algorithms. Not to infer with very less results, we preferred 50, 100, 500, 1000 and 5000 dimensions of the sparsity matrix to trace the effects. The parametric algorithms we used were Linear Discriminant Analysis and Naive Bayes, for non-parametric algorithms we used Decision Trees and Support Vector Machines. These were able to give us results with Precision, Recall and $F_1$-Score for every single dimension. We were able to infer that all the algorithms perform very efficiently at higher representations and it is not that the dimension of the sparsity matrix would interfere with the accuracy of the algorithm. We were also able to derive that Support Vector Machine gave the best results and Linear Discriminant Analysis was a very competitive algorithm, so constraint of parametric and non-parametric does not hold importance, but it is the individual algorithm that is going to provide the best possible results.

\section*{Acknowledgments}
We would genuinely like to thank Andrew Maranhão for imparting the dataset on the Kaggle Platform titled Cyberbullying Classification. We have cited the dataset as prescribed by the author.

\bibliographystyle{unsrt}  
\bibliography{references}

\end{document}